\pdfoutput=1
\documentclass[10pt,journal,compsoc]{IEEEtran}

\ifCLASSOPTIONcompsoc
  \usepackage[nocompress]{cite}
\else
  \usepackage{cite}
  \usepackage{float}
\fi

\usepackage{booktabs}       
\usepackage{makecell}

\usepackage{xcolor}
\usepackage{color}
\usepackage{amsmath}
\usepackage{amsthm}
\usepackage{amssymb}

\usepackage{MnSymbol} 

\usepackage{multirow}
\usepackage{algorithm}
\usepackage{algpseudocode}
\usepackage{graphicx,subfig}
\usepackage{subcaption}

\usepackage{booktabs}       
\usepackage{multirow}       
\usepackage{graphicx}       
\usepackage[table]{xcolor}  
\usepackage{array}          

\usepackage{url}
\usepackage{hyperref}

\definecolor{ccColor}{rgb}{0.788,0.278,0.216}

\begin{document}
%

\title{Generative Models in Decision Making: A Survey}

\author{
    Xinyu Shao, 
    Jianping Zhang, 
    Haozhi Wang, 
    Leo Maxime Brunswic, 
    Kaiwen Zhou, \\
    Jiqian Dong, 
    Kaiyang Guo, 
    Zhitang Chen,
    Jun Wang, 
    Jianye Hao,
    Xiu Li,
    and Yinchuan Li.
    
    \thanks{Corresponding authors: Xiu Li and Yinchuan Li.}
    
    \thanks{X. Shao and X. Li are with the Tsinghua Shenzhen International Graduate School, Tsinghua University, Shenzhen 518055, China. X. Shao is also with the Huawei Noah’s Ark Lab, Shenzhen, China. (e-mail: shaoxy23@mails.tsinghua.edu.cn; li.xiu@sz.tsinghua.edu.cn).}
    
    \thanks{Y. Li, H. Wang, L. M. Brunswic, K. Zhou, J. Dong, K. Guo, Z. Chen, and J. Hao are with the Huawei Noah’s Ark Lab, Shenzhen, China. (e-mail: yinchuan.li.cn@gmail.com)}
    
    \thanks{J. Zhang is with The Chinese University of Hong Kong, Hong Kong SAR, China.}
    
    \thanks{J. Wang is with the Department of Computer Science, University College London, London WC1E 6BT, U.K.}
}

\IEEEtitleabstractindextext{

\begin{abstract}
Generative models have fundamentally reshaped the landscape of decision-making, reframing the problem from pure scalar reward maximization to high-fidelity trajectory generation and distribution matching.
This paradigm shift addresses intrinsic limitations in classical Reinforcement Learning (RL), particularly the limited expressivity of standard unimodal policy distributions in capturing complex, multi-modal behaviors embedded in diverse datasets. 
However, current literature often treats these models as isolated algorithmic improvements, rarely synthesizing them into a single comprehensive framework.
This survey proposes a principled taxonomy grounding generative decision-making within the probabilistic framework of \textbf{Control as Inference}. 
By performing a variational factorization of the trajectory posterior, we conceptualize four distinct functional roles: \textbf{Controllers} for amortized policy inference, \textbf{Modelers} for dynamics priors, \textbf{Optimizers} for iterative trajectory refinement, and \textbf{Evaluators} for trajectory guidance and value assessment.
Unlike existing architecture-centric reviews, this function-centric framework allows us to critically analyze representative generative families across distinct dimensions.
Furthermore, we examine deployment in high-stakes domains, specifically Embodied AI, Autonomous Driving, and AI for Science, highlighting systemic risks such as dynamics hallucination in world models and proxy exploitation.
Finally, we chart the path toward \textbf{Generalist Physical Intelligence}, identifying pivotal challenges in inference efficiency, trustworthiness, and the emergence of Physical Foundation Models.
\end{abstract}

}

\maketitle

\IEEEdisplaynontitleabstractindextext

\IEEEpeerreviewmaketitle

\IEEEraisesectionheading{\section{Introduction}\label{sec:introduction}}

\IEEEPARstart{S}{equential} decision-making has traditionally been dominated by Reinforcement Learning (RL) and optimal control algorithms, which seek to maximize cumulative scalar rewards~\cite{sutton2018reinforcement}. While effective in well-defined simulations, these methods face fundamental bottlenecks when scaled to open-world, high-dimensional tasks.
Although maximum entropy RL methods, such as Soft Actor-Critic (SAC)~\cite{haarnoja2018soft}, attempt to mitigate exploration issues via entropy regularization, they are often constrained by the limited expressivity of parametric distributions (e.g., unimodal Gaussians used in PPO~\cite{schulman2017proximal}).
Consequently, they struggle to capture the complex, multi-modal nature of human behavior found in diverse offline datasets (e.g., D4RL~\cite{fu2020d4rl})~\cite{levine2020offline}, prompting the use of more expressive generative architectures such as diffusion models~\cite{chi2025diffusion}.
Furthermore, the entanglement of dynamics modeling and policy optimization in model-free RL often results in severe sample inefficiency. As the field moves toward generalizing from large-scale datasets and robot foundation models~\cite{brohan2022rt}, the classical trial-and-error paradigm encounters intrinsic limits in both expressivity and robustness, necessitating new approaches that decouple representation learning from behavior synthesis~\cite{ha2018world, lee2024behavior}.

Driven by the success of foundation models in content generation, ranging from DALL-E~\cite{ramesh2021zero, betker2023improving} for imagery to GPT-4~\cite{brown2020language, openai2023gpt} for language, generative models are now reframing decision-making from scalar maximization to \textbf{high-fidelity distribution matching}~\cite{janner2021offline}.
Unlike standard policies that often rely on unimodal or deterministic mappings, models such as Diffusion~\cite{ho2020denoising} and Autoregressive Transformers~\cite{chen2021decision} treat trajectories as first-class data units. This probabilistic perspective offers three distinct advantages:
(1) \textbf{Multimodal Modeling:} They can represent arbitrarily complex, non-parametric distributions, effectively mitigating the mode-collapse issue inherent in classical imitation learning. 
(2) \textbf{Inference-as-Planning:} They transform the hard planning problem into an iterative sampling process, such as denoising in diffusion models, enabling effective search in high-dimensional action spaces. 
(3) \textbf{High-Fidelity Dynamics Modeling:} They act as expressive data-driven simulators that approximate complex physical dynamics, facilitating efficient planning and reducing real-world sample complexity via imagined rollouts.

\begin{figure}[t]
\centering
\includegraphics[width=\columnwidth]{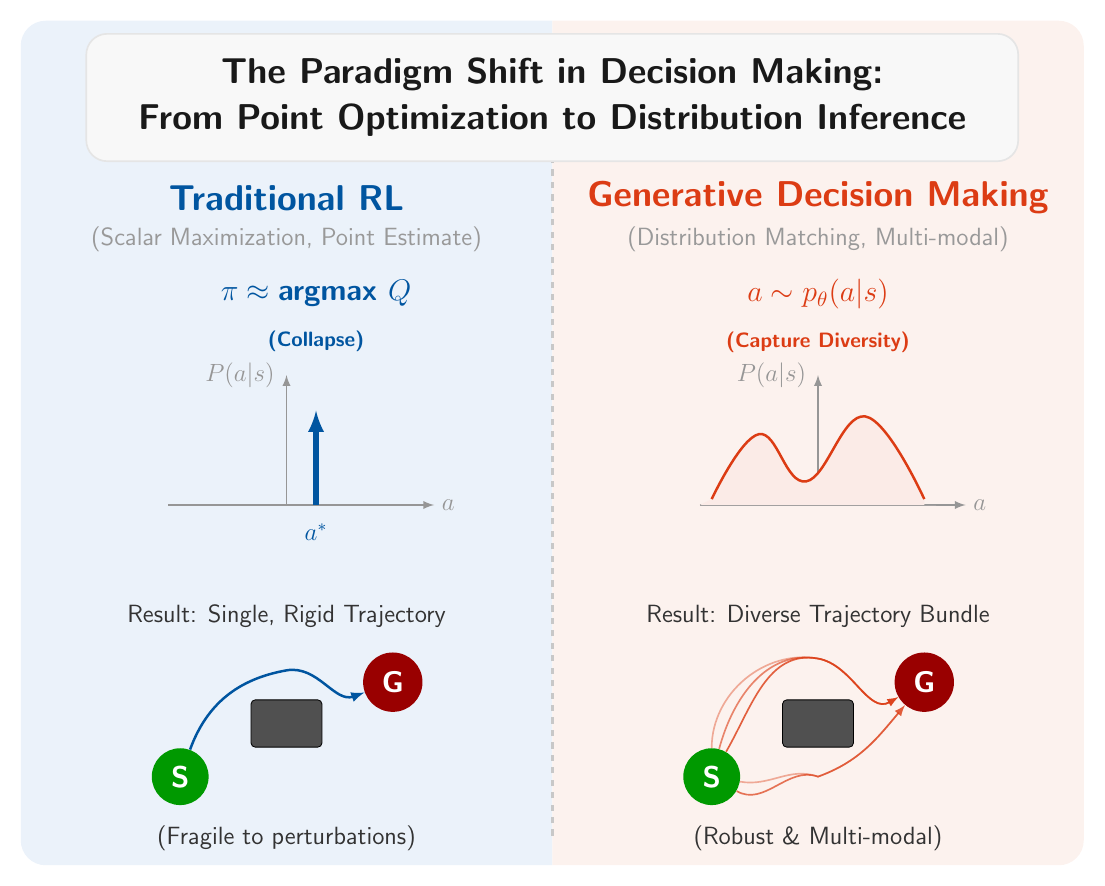}
\caption{The paradigm shift in decision making: from scalar maximization to distribution matching. (Left) Traditional RL typically optimizes for a single optimal policy (point estimate), often leading to mode collapse and rigid behaviors. (Right) Generative Decision Making reframes control as inference, modeling the conditional distribution of optimal trajectories (i.e., the posterior). This approach captures the inherent multi-modal nature of physical behavior data, enabling the generation of diverse, robust, and high-fidelity action sequences.}
\label{fig:concept_shift}
\end{figure}

Despite the surge in research, existing reviews remain fragmented. As detailed in \textbf{Table~\ref{tab:related_surveys}}, prior surveys typically limit their scope to specific architectures, such as Diffusion for RL~\cite{zhu2023diffusion} or Transformers~\cite{li2023survey}, or isolated domains~\cite{gozalo2023chatgpt}. These works often treat generative components as auxiliary modules rather than a core decision-making paradigm. Crucially, they often lack a \textbf{unifying probabilistic framework} to connect diverse mechanisms, from Energy-Based Models to GFlowNets, under a common decision-theoretic lens.

To bridge this gap, this survey proposes a unified taxonomy grounded in the probabilistic framework of \textbf{Control as Inference}. By \textbf{factorizing the trajectory posterior} (detailed in Section~\ref{sec:taxonomy}), we systematically conceptualize four functional roles that cover the complete generative loop: \textbf{Controller}, \textbf{Modeler}, \textbf{Optimizer}, and \textbf{Evaluator}.

The primary contributions of this work are as follows:
\begin{itemize}
    \item \textbf{A Unified, Function-Centric Taxonomy.} We theoretically formulate four roles from the probabilistic factorization of the trajectory posterior. This moves beyond purely architectural categorization to a grounded functional perspective.
    \item \textbf{A Critical Synthesis of Methodologies.} We systematically evaluate representative families across key task dimensions defined in our functional scope, identifying why specific generative mechanisms suit specific decision roles.
    \item \textbf{An Application-Aware Safety Analysis.} We assess real-world applications with a dedicated focus on robustness. This includes identifying systemic risks such as physics hallucination, defined as generating physically implausible transitions in world models, and outlining mitigation strategies for safety.
\end{itemize}

The remainder of this survey is organized as follows. 
Section~\ref{sec:preliminaries} establishes preliminaries. 
Section~\ref{sec:taxonomy} presents our core contribution regarding the theoretical derivation of the unified taxonomy. 
Section~\ref{sec:methodologies} critically analyzes specific algorithms through the lens of this taxonomy. 
Section~\ref{sec:applications} reviews applications with a focus on safety boundaries. 
Finally, Section~\ref{sec:challenges} outlines open challenges and the path toward adaptive and generalizable decision agents.

\begin{table*}[t]
\centering
\caption{\textbf{Comparison of this survey with closely related literature. Prior reviews either focus on specific neural architectures or explore the reverse direction (e.g., RL for Generative AI). In contrast, our work provides a comprehensive, unified perspective grounded in Control as Inference, categorizing all major generative mechanisms by their functional roles.}}
\label{tab:related_surveys}
\renewcommand{\arraystretch}{1.3}
\resizebox{\textwidth}{!}{%
\begin{tabular}{l|>{\raggedright\arraybackslash}p{3.5cm}|>{\raggedright\arraybackslash}p{6.5cm}|c|c}
\toprule
\textbf{Survey (Ref)} & \textbf{Primary Focus / Scope} & \textbf{Taxonomy Basis} & \textbf{Control as Inference Lens} & \textbf{Safety \& Risk Analysis} \\
\midrule
Zhu et al.~\cite{zhu2023diffusion} & Diffusion Models & RL Paradigms (Offline, Online, Multi-task) & $\times$ & Partial \\
Li et al.~\cite{li2023survey} & Transformers & Algorithmic integrations (Model-free/based) & $\times$ & $\times$ \\
Gozalo et al.~\cite{gozalo2023chatgpt} & LLMs / ChatGPT & Application domains (Robotics, Games, etc.) & $\times$ & Partial \\
Cao et al.~\cite{cao2023reinforcement} & RL \textit{for} Generative AI & Alignment techniques (e.g., RLHF, DPO) & $\times$ & $\checkmark$ \\
\rowcolor{blue!5} \textbf{Ours} & \textbf{All Generative Families} (GAN, VAE, Flow, Diff, GFN, AR, EBM) & \textbf{Functional Roles} (Controller, Modeler, Optimizer, Evaluator) & \textbf{$\checkmark$} & \textbf{$\checkmark$ (Systemic Risks)} \\
\bottomrule
\end{tabular}%
}
\end{table*}

\section{Preliminaries}
\label{sec:preliminaries}

In this section, we review the key concepts that help connect probabilistic decision-making with generative Artificial Intelligence (AI). We first introduce the sequential decision problem, framing it from the perspective of \textit{trajectory optimization} to better align with generative modeling frameworks. We then briefly categorize common RL methodologies based on their primary learning objectives (Value vs. Policy) and their use of environmental dynamics (Model-Free vs. Model-Based). Finally, we provide an overview of the generative modeling families that are increasingly being adopted to facilitate decision-making in physical agents.

\subsection{Problem Formulation: From Steps to Trajectories}
\label{subsec:problem_formulation}

\textbf{Markov Decision Processes (MDPs).} 
We formulate the decision-making problem as a Markov Decision Process (MDP)~\cite{bellman1957markovian}, defined by the tuple $\mathcal{M} = (\mathcal{S}, \mathcal{A}, P, R, \gamma, \rho_0)$~\cite{sutton2018reinforcement}.
In partially observable settings (POMDP)~\cite{krishnamurthy2016partially}, states are inferred from observations $o_t$.
Unlike classical controls which focus on instantaneous state $s_t \in \mathcal{S}$ and action $a_t \in \mathcal{A}$, generative decision-making often operates on the level of complete trajectories.
Let $\tau = (s_0, a_0, s_1, a_1, \dots, s_T)$ denote a trajectory sequence of horizon $T$.
The standard objective is to maximize the expected discounted cumulative return $J(\pi) = \mathbb{E}_{\tau \sim \pi}[R(\tau)]$, where $R(\tau) = \sum_{t=0}^T \gamma^t r(s_t, a_t)$.

\textbf{Trajectory Distribution Matching.} 
In the context of generative modeling, we reframe the optimization goal from finding a deterministic policy $\pi^*$ to approximating an optimal trajectory distribution $p^*(\tau)$~\cite{levine2018reinforcement}.
For example, in the offline setting, given a dataset $\mathcal{D} = \{\tau_i\}_{i=1}^N$ drawn from a behavior distribution $\pi_\beta$, the goal is to learn a parameterized policy $\pi_\theta(\tau)$ that minimizes the divergence to the high-reward regions of the data manifold~\cite{levine2020offline, tao2021data}:
\begin{equation}
\label{eq:kl_objective}
    \min_\theta D_{\text{KL}}\left( \pi_\theta(\tau) \,\|\, p(\tau | \mathcal{O}=1) \right),
\end{equation}
where $\mathcal{O}$ denotes the binary event of optimality. This formulation links RL to probabilistic inference, serving as the cornerstone for the unified taxonomy in Section~\ref{sec:taxonomy}.

\subsection{Taxonomy of RL Paradigms}
\label{subsec:rl_taxonomy}

We structure standard RL algorithms along two orthogonal axes: the \textit{learning objective} and the \textit{reliance on dynamics}. Crucially, we highlight how generative approaches address the specific limitations inherent in classical methods.

\subsubsection{Learning Paradigms: Value vs. Policy}

\textbf{Value-Based and Distributional RL.} 
Traditional value-based methods, such as DQN~\cite{mnih2015human} and Fitted Q-iteration~\cite{ernst2005tree}, approximate the expected return $Q^\pi(s,a)$ via Bellman backups ($Q^\pi = \mathcal{B}^\pi Q^\pi$)~\cite{arulkumaran2017deep}. 
However, scalar expectations obscure the multi-modal nature of stochastic environments. 
\textbf{Distributional RL}~\cite{bellemare2017distributional} bridges this gap by modeling the full return distribution $Z^\pi(s,a)$ rather than its mean. 
This shift marks an early precursor to generative decision-making, acknowledging that capturing uncertainty requires estimating densities, not just scalars.

\textbf{Policy-Based and Imitation Learning.} 
Standard policy gradient methods, such as REINFORCE~\cite{sutton1999policy} and PPO~\cite{schulman2017proximal}, typically assume unimodal Gaussian policies $\pi(a|s) = \mathcal{N}(\mu_\theta(s), \Sigma_\theta(s))$. Trust Region methods (TRPO)~\cite{schulman2015trust} stabilize this by constraining the KL divergence between updates.
However, this unimodal assumption is fundamentally limited in open-world settings where valid actions are multi-modal. 
\textbf{Generative Imitation Learning} overcomes this by treating policy learning as conditional density estimation $\pi_\theta(a|s) \approx p_{data}(a|s)$~\cite{ho2016generative}. 
By leveraging rigorous density estimators, these methods can represent arbitrarily complex action distributions, a capability essential for generalist agents~\cite{kuba2021trust, yu2022surprising}.

\subsubsection{Dynamics Modeling: Model-Free vs. Model-Based}

\textbf{Model-Free RL.} 
Agents learn directly from interaction tuples without constructing an environmental surrogate. Advanced Actor-Critic methods like DDPG~\cite{lillicrap2015continuous} and Soft Actor-Critic (SAC)~\cite{haarnoja2018soft} introduce deterministic policies or entropy regularization to improve sample efficiency. While asymptotically optimal, their deployment in physical systems is limited by the high cost of real-world interaction.

\textbf{Model-Based RL (MBRL) and World Models.} 
MBRL approximates transition dynamics $P(s'|s,a)$ to enable planning~\cite{sutton2012introduction}. 
Prior probabilistic approaches often relied on Gaussian ensembles to capture uncertainty (e.g., PETS~\cite{chua2018deep}), which can struggle in high-dimensional visual spaces. 
Modern \textbf{Generative World Models}, derived from the Dyna paradigm~\cite{feinberg2018model}, utilize variational autoencoders (e.g., Dreamer~\cite{hafner2019dream}) or discrete tokens (e.g., IRIS~\cite{micheli2022transformers}) to learn compact latent dynamics. 
This enables agents to perform \textit{planning in imagination}, effectively decoupling physical trial-and-error from cognitive reasoning.

\subsection{Generative Modeling Foundations}
\label{subsec:gen_models}

\begin{figure*}[t]
\centering
\includegraphics[width=0.9\textwidth]{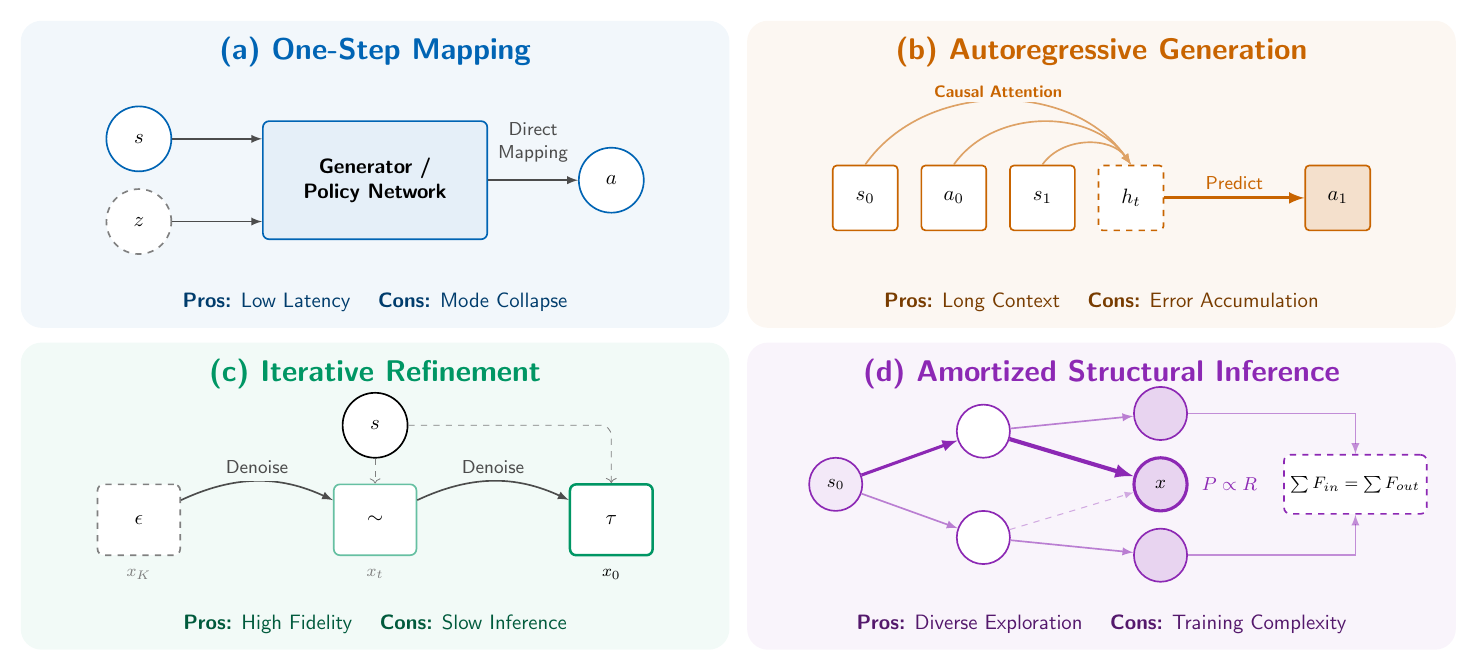}
\caption{\textbf{Schematic comparison of four generative inference mechanisms in decision making.} 
(a) \textbf{One-Step Mapping}: Direct, low-latency projection from state/latent space to actions (e.g., VAEs, GANs). 
(b) \textbf{Autoregressive Generation}: Sequential token prediction utilizing causal attention for long-horizon planning (e.g., AR Models / Transformers). 
(c) \textbf{Iterative Refinement}: Progressive generation via denoising or energy/flow matching, enabling flexible test-time optimization (e.g., Diffusion Models, EBMs, Flow Matching). 
(d) \textbf{Amortized Structural Inference}: Flow-based sampling constructing compositional objects (e.g., molecules or graphs) ensures diverse exploration (e.g., GFlowNets).}
\label{fig:generative_mechanisms}
\end{figure*}

Generative models provide the mathematical machinery to sample from complex distributions $p_{data}(x)$~\cite{goodfellow2020generative, bond2021deep}. In decision-making, $x$ generalizes to actions $a$, states $s$, or trajectories $\tau$. We categorize them into four paradigms based on their inference mechanisms, as illustrated in Figure~\ref{fig:generative_mechanisms}.

\noindent \textbf{One-Step Latent Mapping (VAEs \& GANs).}
Unlike sequence models, these approaches introduce latent variables to capture high-level semantics via a single forward pass. 
\textbf{Variational Autoencoders (VAEs)}~\cite{kingma2013auto, rezende2014stochastic} learn a compressed latent space $\mathcal{Z}$ by maximizing the Evidence Lower Bound (ELBO), which is critical for constructing compact World Models. Extensions like $\beta$-VAE~\cite{higgins2016beta} and VQ-VAE~\cite{van2017neural} further improve disentanglement and discrete representation.
\textbf{Generative Adversarial Networks (GANs)}~\cite{goodfellow2020generative} pioneer implicit sampling via adversarial games. While effective for domain adaptation~\cite{zhu2017unpaired} and imitation~\cite{ho2016generative}, their training instability (mode collapse) often limits their adoption in reliability-critical control compared to likelihood-based methods.

\noindent \textbf{Explicit Sequence Modeling (Autoregressive).}
These models optimize the exact likelihood $p_\theta(x)$ via the chain rule decomposition: $p(x) = \prod_t p(x_t | x_{<t})$~\cite{vaswani2017attention, radford2019language}. 
In decision-making, this mechanism enables casting planning as sequence modeling. Prominent examples include \textbf{Decision Transformer}~\cite{chen2021decision} (for model-free control) and \textbf{Trajectory Transformer}~\cite{janner2021offline} (for model-based planning via beam search). This paradigm allows the massive scaling properties of Large Language Models (LLMs) to be directly transferred to trajectory prediction~\cite{brown2020language, devlin2018bert}.

\noindent \textbf{Iterative Refinement (Diffusion, EBMs \& Flow Matching).}
To model complex continuous distributions without restricting architecture, these models rely on iterative mixing processes. 
Energy-Based Models (EBMs)~\cite{lecun2006tutorial} learn an unnormalized density function via methods like Contrastive Divergence~\cite{hinton2002training}.
Diffusion Models~\cite{sohl2015deep, ho2020denoising, yang2023diffusion} circumvent the intractable normalizing constant by learning the score function $\nabla_x \log p(x)$ via a stochastic differential equation (SDE)~\cite{song2020score}:
\begin{equation}
    dx = [f(x,t) - g^2(t)\nabla_x \log p_t(x)]dt + g(t)dw.
\end{equation}
This iterative process enables \textit{test-time optimization}, allowing planners like Diffuser~\cite{janner2022planning} to refine coarse trajectories into smooth, feasible plans. 
Recently, \textbf{Flow Matching}~\cite{lipman2022flow, liu2023flow} has emerged as a robust alternative, creating straight probability flows that significantly accelerate inference speed for real-time control.

\noindent \textbf{Amortized Structural Inference (GFlowNets).} 
While the above models excel in continuous spaces, decision-making often involves discrete, compositional structures (e.g., molecule graphs). 
We characterize the generation of such objects as \textit{amortized structural inference}, where the high cost of exploring complex discrete distributions is amortized into rapid, sequential sampling steps. As a prominent example, \textbf{Generative Flow Networks (GFlowNets)}~\cite{bengio2021gflownet, bengio2021flow} achieve this by treating policy learning as flow matching on a directed acyclic graph (DAG). Unlike RL which maximizes reward, GFlowNets aim to sample objects $x$ proportional to their reward $R(x)$, satisfying the flow consistency constraint:
\begin{equation}
    \sum_{s' \to s} F(s' \to s) = \sum_{s \to s''} F(s \to s'').
\end{equation}
This property makes them uniquely suited for \textbf{diverse exploration} in vast combinatorial spaces~\cite{malkin2022trajectory, pan2023better}.

\section{Unified Taxonomy: Control as Inference}
\label{sec:taxonomy}

Current literature predominantly categorizes methods by architecture (e.g., Diffusion vs. Transformer). However, this \textit{architecture-centric} view is increasingly insufficient because structural equivalence does not imply functional equivalence. For instance, a Transformer can serve as a \textbf{Controller} (Decision Transformer) or a \textbf{Modeler} (IRIS). To navigate this complexity, a taxonomy grounded in \textit{decision-theoretic roles} rather than \textit{backbone networks} is essential.

We propose a unified framework rooted in \textbf{Control as Inference}~\cite{levine2018reinforcement}. We posit that generative models are naturally suited for this paradigm, acting as powerful \textbf{approximate inference engines} to solve the intractable trajectory posterior. By formally factorizing this objective, we derive four canonical roles: \textbf{Controller}, \textbf{Modeler}, \textbf{Evaluator}, and \textbf{Optimizer}. By decoupling architecture from functional purpose, this taxonomy allows us to systematically analyze how different generative mechanisms address specific decision-making bottlenecks. We detail the mathematical derivation of these roles below.

\subsection{Theoretical Foundation: Control as Inference}
\label{subsec:theory}

Consider a trajectory $\tau = (s_0, a_0, \dots, s_T, a_T)$. To frame reward maximization as a probabilistic inference problem, we introduce a binary optimality variable $\mathcal{O}_t$. Rather than treating raw rewards as probabilities, we define the \textbf{unnormalized likelihood} of a specific step being optimal as $p(\mathcal{O}_t=1 | s_t, a_t) \propto \exp(r(s_t, a_t))$. This exponential transformation elegantly maps arbitrary scalar rewards to non-negative potential values. Consequently, the fundamental goal of decision-making is to infer the posterior distribution of trajectories conditioned on the optimality of all steps: $p(\tau | \mathcal{O}_{1:T}=1)$.

Using Bayes' rule and the Markov property, this posterior factorizes as:
\begin{align}
\label{eq:factorization}
    p(\tau | \mathcal{O}) & \propto \underbrace{p(\tau)}_{\text{Trajectory Prior}} \cdot \underbrace{p(\mathcal{O} | \tau)}_{\substack{\text{Optimality} \\ \text{Likelihood}}} \\
    & = \rho_0(s_0) \bigg( \prod_{t=0}^{T-1} \underbrace{p(s_{t+1} | s_t, a_t)}_{\substack{\text{Dynamics} \\ \text{(Modeler)}}} \underbrace{\pi(a_t | s_t)}_{\substack{\text{Policy} \\ \text{(Controller)}}} \bigg) \cdot \underbrace{\exp(R(\tau))}_{\substack{\text{Value} \\ \text{(Evaluator)}}}. \nonumber
\end{align}

This factorization (Eq.~\ref{eq:factorization}) reveals the fundamental components of generative decision-making, demonstrating that these four roles are necessary and sufficient to cover the entire inference process:
\begin{itemize}
    \item \textbf{Controller ($\pi$):} The policy prior $\pi(a|s)$, providing the proposal distribution for actions.
    \item \textbf{Modeler ($P$):} The transition dynamics $p(s'|s,a)$, defining the physical laws of the environment.
    \item \textbf{Evaluator ($R$):} The optimality likelihood $\exp(R(\tau))$, representing goal or constraint satisfaction.
    \item \textbf{Optimizer (Inference):} The algorithmic mechanism (e.g., variational inference or iterative sampling) used to approximate the intractable posterior $p(\tau|\mathcal{O})$.
\end{itemize}

\subsection{Functional Roles and Scope Definition}
\label{subsec:roles}

Based on the variational derivation, we define the precise scope for each role. \textbf{Table~\ref{tab:scope_box}} serves as the Scope Box for our taxonomy, mapping each role to its theoretical inputs, outputs, and the specific assumptions generative models operate under.

\begin{table*}[t]
\centering
\caption{\textbf{The Scope Box: Definitions and Interfaces of Generative Roles.} This table formally maps each functional role to its mathematical equivalent in the Control as Inference framework, explicitly defining the boundaries, inputs, outputs, and underlying assumptions.}
\label{tab:scope_box}
\renewcommand{\arraystretch}{1.3}
\resizebox{\textwidth}{!}{%
\begin{tabular}{l|l|l|l|l|l}
\toprule
\textbf{Role} & \textbf{Theoretical Equivalent} & \textbf{Input Space} & \textbf{Output / Target} & \textbf{Assumption} & \textbf{Core Generative Task} \\
\midrule
\textbf{Controller} & Policy Prior $\pi(a|s)$ & State $s$ (or History $h$) & Action $a$ & Markovian / Auto-regressive & \textbf{Amortized Sampling:} Instantly generating optimal actions from states. \\
\textbf{Modeler} & Dynamics $p(s'|s,a)$ & State $s$, Action $a$ & Next State $s'$ / Reward & Stationary Dynamics & \textbf{Density Estimation:} Simulating environment transitions and rollouts. \\
\textbf{Evaluator} & Likelihood $p(\mathcal{O}|\tau)$ & State $s$, Action $a$, or $\tau$ & Scalar Score / Safety Flag & Tractable Likelihood / Density & \textbf{Guidance \& Verification:} Providing gradients or rejecting unsafe samples. \\
\textbf{Optimizer} & Posterior $q \approx p(\tau|\mathcal{O})$ & State $s_0$, Goal $g$, Noise $\epsilon$ & Trajectory $\tau$ & Known Reward / Differentiable & \textbf{Iterative Planning:} Refining sequences via gradients or sampling (e.g., denoising). \\
\bottomrule
\end{tabular}%
}
\end{table*}

\vspace{6pt}
\noindent \textbf{Controller (The Amortized Inference).}
Generative models in this role perform \textit{amortized inference}: they learn a parametric map $\pi_\theta$ to directly approximate the optimal posterior. Unlike standard RL policies which are often deterministic or unimodal Gaussian, generative controllers, such as Diffusion Policies~\cite{chi2025diffusion}, can represent highly \textbf{multi-modal action distributions}. This is crucial for offline imitation learning, where human demonstrations are naturally diverse and multi-modal.

\vspace{6pt}
\noindent \textbf{Modeler (The Dynamics Prior).}
These models approximate the environment dynamics $p(s'|s,a)$. In the context of this factorization, they serve as the prior that constrains the optimization to physically plausible trajectories. A generative modeler acts as a World Model~\cite{ha2018world}, allowing the agent to dream potential futures. Crucially, generative models enable \textbf{high-fidelity simulation} in complex domains (e.g., video prediction) that are intractable for traditional Gaussian dynamics models.

\vspace{6pt}
\noindent \textbf{Evaluator (The Likelihood Estimator).}
The Evaluator approximates the optimality likelihood $p(\mathcal{O}|\tau)$. While standard RL relies on scalar Value Functions, generative approaches often employ Energy-Based Models (EBMs) or Discriminators. Their key advantage is providing \textbf{dense gradient signals} ($\nabla_\tau \log p(\mathcal{O}|\tau)$) rather than sparse rewards, guiding the optimizer toward high-reward regions via differentiable manifolds. Furthermore, in safety-critical systems, Evaluators function as \textbf{Safety Guards}, filtering out generated trajectories that violate learned constraints via rejection sampling.

\vspace{6pt}
\noindent \textbf{Optimizer (The Iterative Inference).}
The Optimizer is the mechanism that performs the maximization of the objective. We classify methods as Optimizers when the generative model is used to define the \textit{search process} itself, for example in Diffuser~\cite{janner2022planning}. Here, trajectory planning is treated as a generative in-painting problem. These methods perform computation-heavy \textbf{iterative inference} at test time (e.g., via reverse diffusion) to refine trajectories, offering stronger mode-seeking capabilities and long-horizon consistency than single-step policies.

\vspace{6pt}
\noindent \textbf{Remark on Functional Overlap.}
In advanced architectures, boundaries between roles can blur. We categorize such methods based on their \textit{inference-time behavior}. 
For instance, Diffuser~\cite{janner2022planning} models the joint distribution $p(\tau)$. Although it conceptually acts as a policy, its inference involves an iterative denoising process guided by rewards. Therefore, we classify it primarily as an \textbf{Optimizer} because the core contribution is the planning-as-sampling mechanism. 
Conversely, Decision Transformer~\cite{chen2021decision} performs autoregressive sequence modeling. While it models the joint distribution $p(\tau)$, its inference is a direct, causal token generation. Consequently, we classify it as a Controller performing amortized policy inference.

\begin{figure}[t]
    \centering
    \includegraphics[width=\linewidth]{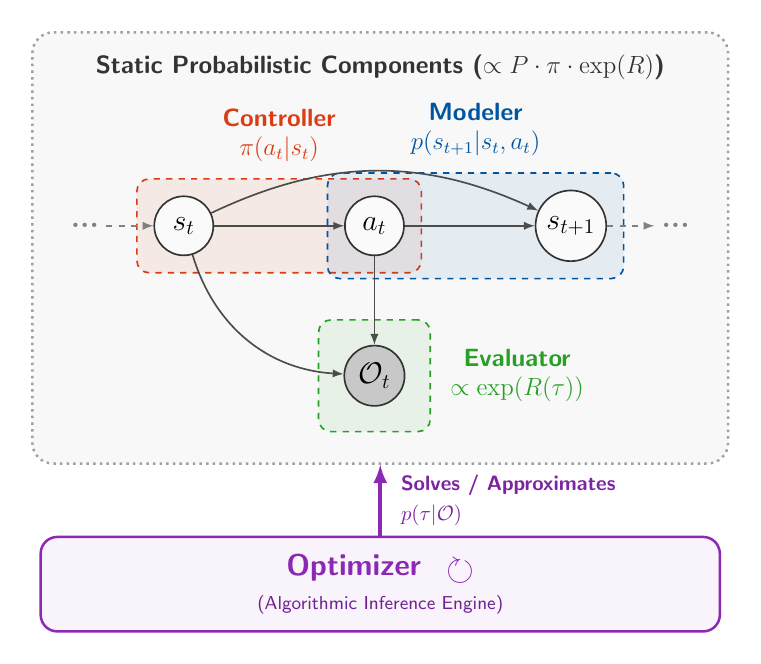}
    \caption{The Unified Taxonomy of Generative Decision Making based on Control as Inference. Following Equation~\ref{eq:factorization}, we decompose the decision-making process into four probabilistic components: (1) \textbf{Controller}: The policy prior $\pi(a_t|s_t)$ that proposes actions; (2) \textbf{Modeler}: The dynamics model $p(s_{t+1}|s_t, a_t)$ that predicts future states; (3) \textbf{Evaluator}: The likelihood function $p(\mathcal{O}|\tau)$ that evaluates optimality; and (4) \textbf{Optimizer}: The inference mechanism that solves for the posterior $p(\tau|\mathcal{O})$ to determine optimal actions.}
    \label{fig:taxonomy}
\end{figure}

\subsection{Task Dimensions}
\label{subsec:task_dimensions}

Beyond functional roles, we categorize methodologies according to four key decision-making dimensions. These dimensions dictate the specific constraints and determine which generative advantages are most critical.

\vspace{6pt}
\noindent \textbf{Online vs. Offline Settings.}
\textit{Offline} settings require models to handle distribution shifts and stitch suboptimal sub-trajectories; here, generative models excel due to their superior distribution matching capabilities. \textit{Online} settings demand efficient exploration and real-time inference, often requiring lighter-weight generative models or distillation techniques to minimize latency.

\vspace{6pt}
\noindent \textbf{State and Action Modality.}
Tasks range from low-dimensional proprioceptive states to high-dimensional visual observations (POMDPs). Generative models, such as Latent Diffusion or VAEs, are particularly adept at handling partial observability and visual complexity by learning compact belief states in latent space, enabling planning directly on pixels.

\vspace{6pt}
\noindent \textbf{Learning Signal (Reward vs. Preference).}
While traditional RL maximizes scalar rewards, recent trends, such as RLHF, utilize preference rankings. Generative Evaluators (Reward Models) play a central role here, bridging static preference datasets with policy optimization by learning a differentiable reward landscape that guides the agent.

\vspace{6pt}
\noindent \textbf{Single vs. Multi-Agent Contexts.}
In multi-agent scenarios, the goal extends to coordinating joint actions or modeling opponents. Generative modeling provides a distinct advantage here by approximating complex joint equilibrium distributions, allowing agents to sample diverse and consistent joint strategies.

To ground this taxonomy in existing literature, we provide a comprehensive catalog of representative algorithms in Table~\ref{tab:comprehensive_catalog_part1} (split into Part 1 and Part 2 due to space constraints). These tables map state-of-the-art methods across various application domains to their corresponding generative families and functional roles, serving as a roadmap for the subsequent methodological analysis.

\begin{table*}[t]
    \centering
    \caption{A Comprehensive Catalog of Generative Decision-Making Algorithms (Part 1 of 2). Methods are categorized by their primary \textbf{Functional Role} (as defined in Section~\ref{sec:taxonomy}) and Generative Family. Key task dimensions (Online/Offline, Domain) are highlighted.}
    \label{tab:comprehensive_catalog_part1}
    \renewcommand{\arraystretch}{1.0} 
    \setlength{\tabcolsep}{3pt}
    \footnotesize 
    \begin{tabular*}{\textwidth}{@{\extracolsep{\fill}} l|l|l|l|l|l @{}}
    \toprule
    \rowcolor{gray!15} \textbf{Role} & \textbf{Algorithm (Ref)} & \textbf{Gen. Family} & \textbf{Backbone / Structure} & \textbf{Setting (RL Paradigm)} & \textbf{Application Domain} \\
    \midrule

    \multirow{43}{*}{\rotatebox{90}{\textbf{CONTROLLER (Policy)}}} 
    & Soft Q-learning~\cite{haarnoja2017reinforcement} & EBM & Energy Function & Online RL & Robot Control \\
    & EBIL~\cite{liu2021energy} & EBM & Energy Function & Imitation Learning & Robot Control \\
    \cmidrule{2-6}
    & GAIL~\cite{ho2016generative} & GAN & Generator-Discriminator & Imitation Learning & Robot Control \\
    & InfoGAIL~\cite{li2017infogail} & GAN & InfoGAN & Imitation Learning & Robot Control \\
    & MGAIL~\cite{baram2017end} & GAN & GAN & Imitation Learning & Robot Control \\
    & FAGIL~\cite{geiger2022fail} & GAN & Wasserstein GAN & Imitation Learning & Robot Control \\
    & WGAIL~\cite{wang2021learning} & GAN & GAN & Imitation Learning & Robot Control \\
    & IC-GAIL~\cite{wu2019imitation} & GAN & GAN & Imitation Learning & Robot Control \\
    & AIRL~\cite{fu2018learning} & GAN & GAN (Inv. RL) & Imitation Learning & Robot Control \\
    & AugAIRL~\cite{wang2021decision} & GAN & GAN & Imitation Learning & Robot Control \\
    & WAIL~\cite{xiao2019wasserstein} & GAN & Wasserstein GAN & Imitation Learning & Robot Control \\
    & MAIRL~\cite{sun2021adversarial} & GAN & GAN & Imitation Learning & Robot Control \\
    \cmidrule{2-6}
    & GTI~\cite{mandlekar2020learning} & VAE & CVAE & Imitation Learning & Robot Control \\
    & HULC~\cite{mees2022matters} & VAE & Seq2Seq CVAE & Imitation Learning & Robot Control \\
    & Play-LMP~\cite{lynch2020learning} & VAE & Seq2Seq CVAE & Imitation Learning & Robot Control \\
    & TACO-RL~\cite{rosete2023latent} & VAE & Seq2Seq CVAE & Imitation Learning & Robot Control \\
    & OPAL~\cite{ajay2020opal} & VAE & $\beta$-VAE & Offline RL & Structural Gen. \\
    & MaskDP~\cite{liu2022masked} & VAE & Masked Autoencoder & Offline RL & Robot Control \\
    \cmidrule{2-6}
    & NF Policy~\cite{ward2019improving} & Flow & Coupling Flow & Offline RL & Robot Control \\
    & CNF~\cite{akimov2022let} & Flow & Autoregressive Flow & Offline RL & Autonomous Driving \\
    & Guided Flows~\cite{zheng2023guided} & Flow & Continuous Flow & Offline RL & Optimization \\
    \cmidrule{2-6}
    & Pearce et al.~\cite{pearce2022imitating} & Diffusion & DDPM & Imitation Learning & Robot Control \\
    & Diffusion Policy~\cite{chi2025diffusion} & Diffusion & DDPM (U-Net) & Imitation (Robotics) & Robot Control \\
    & Diffuser~\cite{janner2022planning} & Diffusion & DDPM (Trajectory) & Offline RL & Structural Gen. \\
    & Decision Diffuser~\cite{ajay2023is} & Diffusion & DDPM (Inv. Dynamics) & Offline RL & Structural Gen. \\
    & Decision Stacks~\cite{zhao2024decision} & Diffusion & DDPM & Offline RL & Structural Gen. \\
    & Diffusion-QL~\cite{wang2023diffusion} & Diffusion & DDPM & Offline RL & Robot Control \\
    & SfBC~\cite{chen2023offline} & Diffusion & DDPM & Offline RL & Structural Gen. \\
    & AdaptDiffuser~\cite{liang2023adaptdiffuser} & Diffusion & DDPM & Robotics & Robot Control \\
    & DIPO~\cite{yang2023policy} & Diffusion & DDPM & Online RL & Robot Control \\
    & UniPi~\cite{du2024learning} & Diffusion & DDPM & Offline RL & Robot Control \\
    \cmidrule{2-6}
    & GFlowNets~\cite{bengio2021gflownet} & GFlowNet & Trajectory Flow & Offline/Online RL & Structural Gen. \\
    & Stochastic GFN~\cite{pan2023stochastic} & GFlowNet & GFlowNet & Offline RL & Structural Gen. \\
    & GAFlowNets~\cite{pan2022generative} & GFlowNet & GFlowNet & Offline RL & Structural Gen. \\
    & AFlowNets~\cite{jiralerspong2023expected} & GFlowNet & GFlowNet & Offline RL & Structural Gen. \\
    & CFlowNets~\cite{li2023cflownets} & CFlowNet & Continuous Flow & Online RL & Structural Gen. \\
    \cmidrule{2-6}
    & Decision Trans.~\cite{chen2021decision} & Autoregressive & GPT (Decoder) & Offline RL & Robot Control \\
    & Trajectory Trans.~\cite{janner2021offline} & Autoregressive & GPT (Decoder) & Offline RL & Robot Control \\
    & Gato~\cite{reed2022generalist} & Autoregressive & Multi-modal GPT & Offline RL & Generalist Agent \\
    & Multi-Game DT~\cite{lee2022multigame} & Autoregressive & GPT (Decoder) & Offline RL & Games \\
    & Online DT~\cite{zheng2022online} & Autoregressive & GPT (Decoder) & Online RL & Driving \\
    & PEDA~\cite{zhu2023scaling} & Autoregressive & GPT (Decoder) & Offline RL & Robot Control \\
    & BooT~\cite{wang2022bootstrapped} & Autoregressive & GPT (Decoder) & Offline RL & Structural Gen. \\
    \bottomrule
    \end{tabular*}
\end{table*}

\addtocounter{table}{-1} 
\begin{table*}[t]
    \centering
    \caption{A Comprehensive Catalog of Generative Decision-Making Algorithms (Part 2 of 2 - Continued). This section covers the Modeler, Optimizer, and Evaluator roles.}
    \label{tab:comprehensive_catalog_part2}
    \renewcommand{\arraystretch}{1.0}
    \setlength{\tabcolsep}{3pt}
    \footnotesize 
    \begin{tabular*}{\textwidth}{@{\extracolsep{\fill}} l|l|l|l|l|l @{}}
    \toprule
    \rowcolor{gray!15} \textbf{Role} & \textbf{Algorithm (Ref)} & \textbf{Gen. Family} & \textbf{Backbone / Structure} & \textbf{Setting (RL Paradigm)} & \textbf{Application Domain} \\
    \midrule

    \multirow{13}{*}{\rotatebox{90}{\textbf{MODELER}}} 
    & BM~\cite{ackley1985learning} & EBM & Boltzmann Machine & Generation & Structural Gen. \\
    & DEBMs~\cite{haarnoja2017reinforcement} & EBM & EBM & Online RL & Robot Control \\
    & SGMs~\cite{song2019generative} & EBM & Score Model & Generation & Structural Gen. \\
    \cmidrule{2-6}
    & EGAN~\cite{huang2017enhanced} & GAN & GAN & Online RL & Structural Gen. \\
    & S2P~\cite{cho2022s2p} & GAN & GAN & Offline RL & Robot Control \\
    \cmidrule{2-6}
    & Han \& Kim~\cite{han2022selective} & VAE & VAE & Offline RL & Structural Gen. \\
    \cmidrule{2-6}
    & NICE~\cite{dinh2014nice} & Flow & Coupling Flow & Generation & Optimization \\
    & Rezende et al.~\cite{rezende2015variational} & Flow & Norm. Flow & Generation & Optimization \\
    \cmidrule{2-6}
    & MTDiff~\cite{he2024diffusion} & Diffusion & DDPM & Offline RL & Robot Control \\
    & GenAug~\cite{chen2023genaug} & Diffusion & Latent Diffusion & Robotics & Sim Augmentation \\
    & SynthER~\cite{lu2024synthetic} & Diffusion & EDM & Offline/Online RL & Robot Control \\
    \cmidrule{2-6}
    & ALPINE~\cite{wang2024alpine} & Autoregressive & Transformer & Online RL & Optimization \\
    & ARP~\cite{korenkevych2019autoregressive} & Autoregressive & Transformer & Online RL & Games \\
    \midrule

    \multirow{16}{*}{\rotatebox{90}{\textbf{OPTIMIZER}}} 
    & SO-EBM~\cite{kong2022end} & EBM & Energy Landscape & Optimization & Optimization \\
    & pcEBM~\cite{tagasovska2022pareto} & EBM & Pareto Front & Generation & Structural Gen. \\
    & CF-EBM~\cite{zhao2020learning} & EBM & Energy & Generation & Structural Gen. \\
    \cmidrule{2-6}
    & He et al.~\cite{he2020evolutionary} & GAN & GAN & Generation & Optimization \\
    & DCGANs~\cite{sim2021gans} & GAN & DCGAN & Generation & Optimization \\
    & C-GANs~\cite{kalehbasti2021augmenting} & GAN & C-GAN & Generation & Optimization \\
    \cmidrule{2-6}
    & CVAE-Opt~\cite{hottung2021learning} & VAE & CVAE & Optimization & Optimization \\
    & CageBO~\cite{xing2023bayesian} & VAE & CVAE & Optimization & Optimization \\
    \cmidrule{2-6}
    & Gabrié et al.~\cite{gabrie2022adaptive} & Flow & Norm. Flow & Generation & Optimization \\
    \cmidrule{2-6}
    & DDOM~\cite{krishnamoorthy2023diffusion} & Diffusion & DDPM & Optimization & Optimization \\
    & Li et al.~\cite{li2024diffusion} & Diffusion & DDIM & Optimization & Optimization \\
    & DiffOPT~\cite{kong2024diffusion} & Diffusion & DDIM & Optimization & Optimization \\
    \cmidrule{2-6}
    & GFACS~\cite{kim2024ant} & GFlowNet & GFlowNet & Generation & Optimization \\
    & MOGFNs~\cite{jain2023multi} & GFlowNet & Cond. Flow & Generation & Optimization \\
    \cmidrule{2-6}
    & BONET~\cite{mashkaria2023generative} & Autoregressive & Decoder Only & Optimization & Optimization \\
    & TNP~\cite{nguyen2022transformer} & Autoregressive & Encoder-Decoder & Optimization & Optimization \\
    \midrule

    \multirow{6}{*}{\rotatebox{90}{\textbf{EVALUATOR}}} 
    & EBIL (Reward)~\cite{liu2021energy} & EBM & Energy Function & Inverse RL & Reward Modeling \\
    & DEBMs (Cost)~\cite{haarnoja2017reinforcement} & EBM & Energy Function & Online RL & Soft Constraints \\
    \cmidrule{2-6}
    & GAIL (Discrim.)~\cite{ho2016generative} & GAN & Discriminator & Inverse RL & Surrogate Reward \\
    & AIRL (Reward)~\cite{fu2018learning} & GAN & Discriminator & Inverse RL & Reward Learning \\
    \cmidrule{2-6}
    & PlanCP~\cite{sun2024conformal} & Statistical & Conformal Prediction & Safety Guard & Autonomous Driving \\
    & KnowNo~\cite{ren2023knowno} & Statistical & Conformal Prediction & Safety Guard & Robot Control \\
    \bottomrule
    \end{tabular*}
\end{table*}

\begin{figure*}[t]
\centering
\includegraphics[width=\textwidth]{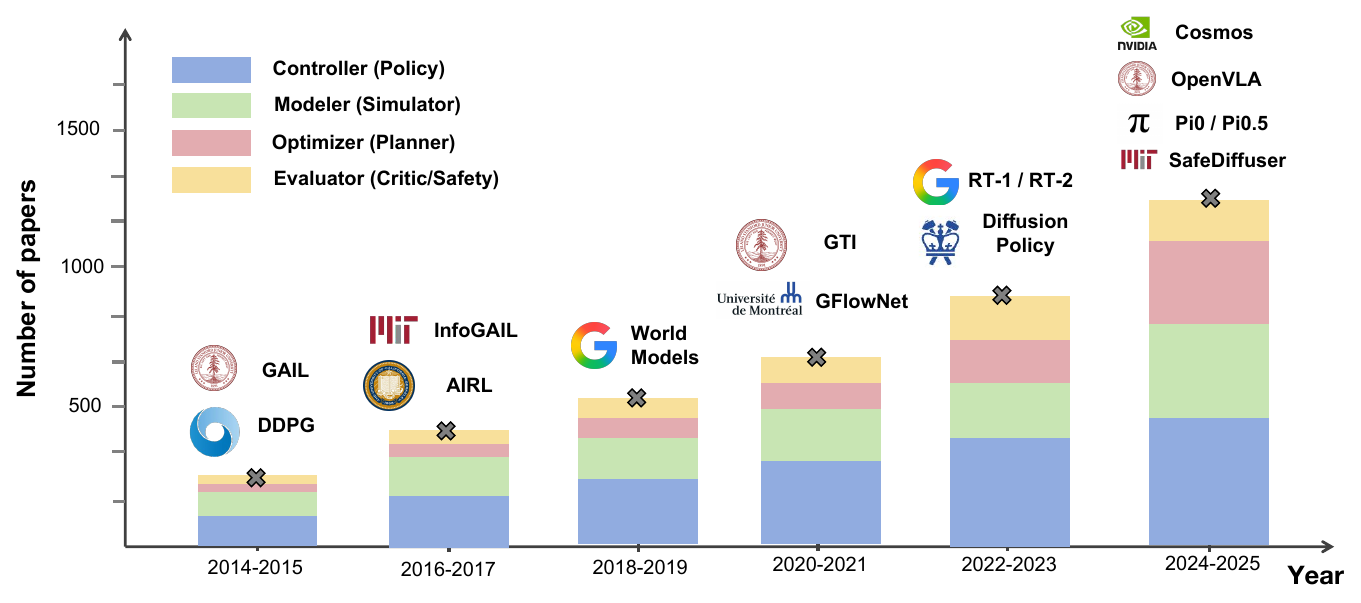}
\caption{\textbf{The evolutionary trajectory of generative models in physical decision-making.} The bar chart illustrates the exponential growth in publications and the paradigm shift across our proposed functional roles (Controller, Modeler, Optimizer, Evaluator). While early research heavily focused on amortized policy inference (Controllers, e.g., DDPG), recent years have witnessed a massive surge in generative environment simulators (Modelers, e.g., Cosmos) and iterative planning mechanisms (Optimizers, e.g., Diffuser). Highlighted works represent seminal milestones bridging generative AI and physical intelligence. Publication statistics were aggregated from Google Scholar using targeted keywords to reflect the community's shifting research priorities and validate our proposed taxonomy.}
\label{fig:trend_bar_chart}
\end{figure*}

\section{Methodologies: A Critical Analysis}
\label{sec:methodologies}

As visually evidenced by the historical trajectory in \textbf{Figure~\ref{fig:trend_bar_chart}}, the landscape of generative decision-making has undergone a profound evolution over the past decade. While early applications were heavily dominated by \textit{Controllers} designed for direct policy imitation, the advent of scalable sequence modeling and iterative refinement has recently catalyzed a massive surge in \textit{Modelers} (generative world simulators) and \textit{Optimizers} (inference-based planners). 

\begin{figure}[t]
\centering
\includegraphics[width=\columnwidth]{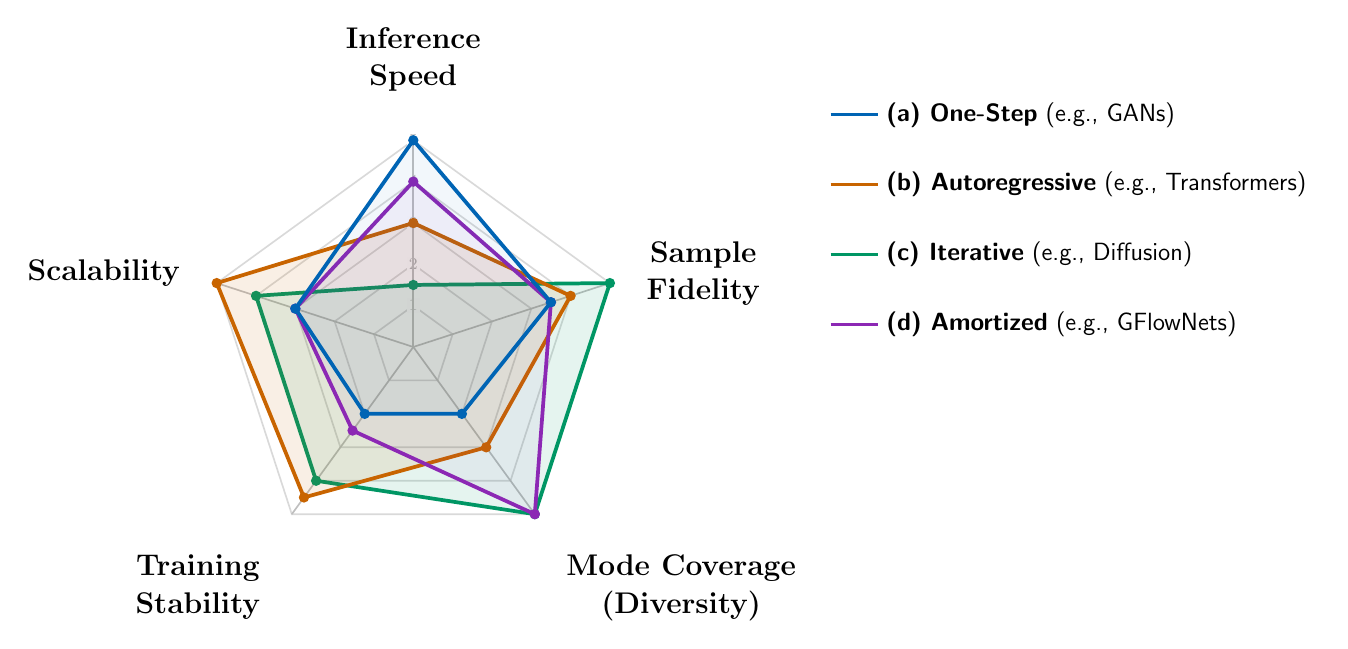} 
\caption{\textbf{Qualitative trade-off analysis of four generative paradigms across five critical dimensions.} 
(a) \textbf{One-Step Mapping} (Blue) excels in speed but lacks diversity.
(b) \textbf{Autoregressive} (Orange) offers extreme scalability but suffers from error accumulation.
(c) \textbf{Iterative Refinement} (Green) achieves high fidelity and mode coverage at the cost of inference speed.
(d) \textbf{Amortized Structural Inference} (Purple) specializes in diversity for discrete structures but faces training stability challenges.}
\label{fig:tradeoff_radar}
\end{figure}

Building upon this historical context, this section critically synthesizes how different generative mechanisms fulfill the four functional roles defined in our taxonomy. Rather than exhaustively listing every algorithm, we categorize methods by their \textbf{inference mechanisms} and analyze their comparative advantages and trade-offs in decision-making contexts. A comprehensive synthesis of these mechanisms is detailed in \textbf{Table~\ref{tab:mechanism_compare}}, while \textbf{Figure~\ref{fig:tradeoff_radar}} visually maps these trade-offs across five critical performance dimensions to serve as a rapid navigational guide.

\begin{table*}[t]
\centering
\caption{\textbf{Comparative Analysis of Generative Mechanisms across Functional Roles.} This table synthesizes the core advantages, inference trade-offs, and primary utilities of different generative paradigms when acting as Controllers, Modelers, Optimizers, and Evaluators, providing a when-to-use rule of thumb.}
\label{tab:mechanism_compare}
\renewcommand{\arraystretch}{1.2}
\resizebox{\textwidth}{!}{%
\begin{tabular}{l|l|l|l|l}
\toprule
\textbf{Functional Role} & \textbf{Implementation Paradigm} & \textbf{Core Advantage / Output} & \textbf{Key Trade-off / Limitation} & \textbf{Primary Utility} \\
\midrule
\multirow{4}{*}{\textbf{Controller (Policy)}}
& One-Step Mapping (GAN / VAE) & Fast (1-step) Inference & Mode collapse / Blurry actions & High-frequency reactive control \\
& Iterative Refinement (Diffusion / Flow) & High Mode Coverage & \textbf{High inference latency} & High-fidelity offline imitation \\
& Sequence Modeling (Autoregressive) & Extreme Scalability & Error compounding over time & Generalist multi-task agents \\
\midrule
\multirow{3}{*}{\textbf{Modeler (Simulator)}}
& Latent Dynamics (VAE / RSSM) & \textbf{Fast} state transitions & Medium fidelity (Posterior collapse) & Online Model Predictive Control \\
& Token Prediction (Autoregressive) & Long-horizon consistency & Quantization errors & Scalable general world models \\
& Pixel Synthesis (Diffusion / GAN) & \textbf{Very High} visual fidelity & Slow generation speed & Offline synthetic data augmentation \\
\midrule
\multirow{3}{*}{\textbf{Optimizer (Planner)}}
& Trajectory In-painting (Diffusion) & Long-horizon consistency & Computationally heavy at test-time & Continuous control / Navigation \\
& Proportional Sampling (GFlowNet) & Mode Diversity & Complex training stability & Discrete / Combinatorial Design \\
& Latent Space Search (VAE / GAN) & Landscape Smoothing & Strictly bound to latent quality & Black-box Policy Search \\
\midrule
\multirow{3}{*}{\textbf{Evaluator (Critic)}}
& Energy Guidance (EBM) & Energy Gradient $\nabla E$ & Requires expensive MCMC sampling & Differentiable Planning / Constraints \\
& Adversarial Scoring (GAN) & Real/Fake Probability & Adversarial game non-stationarity & Inverse RL / Surrogate Reward \\
& Density Monitoring (Flow / VAE) & Exact Likelihood $\log p(x)$ & Compute-heavy for strict bounds & \textbf{Safety Guard} / OOD Detection \\
\bottomrule
\end{tabular}%
}
\end{table*}

\subsection{Generative Models as Controller (The Policy)}
\label{subsec:controller}

Recall from Eq.~(\ref{eq:factorization}) that the Controller corresponds to the policy prior $\pi(a_t|s_t)$, whose theoretical objective is to approximate the optimal trajectory distribution. The Controller role maps states $s_t$ (or history $h_t$) to actions $a_t$. While standard RL relies on unimodal Gaussian assumptions ($\pi(a|s) = \mathcal{N}(\mu, \sigma)$), generative controllers leverage advanced density estimation to capture complex, multi-modal policies~\cite{hazan2020nonstochastic, chen2021black}. This is particularly critical in \textbf{Imitation Learning}, where human demonstrators often exhibit diverse strategies for the same task.

\vspace{6pt}
\noindent \textbf{One-Step Mappings: GANs and VAEs.}
Early approaches focused on direct, single-step inference. \textbf{GAN-based controllers}, such as GAIL~\cite{ho2016generative} and its variants (e.g., InfoGAIL~\cite{li2017infogail}, FAGIL~\cite{geiger2022fail}), employ a discriminator to force the policy to match expert occupancy measures. Extensions like AIRL~\cite{fu2018learning} and AugAIRL~\cite{wang2021decision} further frame this as adversarial reward learning, while others integrate optimal transport (WAIL~\cite{xiao2019wasserstein}) or model-based differentiable optimization (MAIRL~\cite{sun2021adversarial}). While offering ultra-fast inference suitable for real-time control, they notoriously suffer from \textbf{mode collapse}, often dropping diverse strategies to focus on a single dominant mode.

Conversely, \textbf{VAE-based policies} utilize latent spaces to structure exploration. Examples include Play-LMP~\cite{lynch2020learning} for learning from play and TACO-RL~\cite{rosete2023latent} for long-horizon hierarchical control. For primitive learning, OPAL~\cite{ajay2020opal} extracts continuous latent behaviors, while \textbf{recent advancements}~\cite{lee2024behavior} extend this paradigm by employing discrete latent actions to further enhance behavior generation robustness. Approaches like MaskDP~\cite{liu2022masked} and HULC~\cite{mees2022matters} further enhance this by integrating masking or hierarchical decomposition. However, the MSE-based reconstruction loss typically leads to averaged or blurry actions, which lacks the precision required for fine-grained manipulation tasks.

\vspace{6pt}
\noindent \textbf{Iterative Refinement: Diffusion and Flow Policies.}
Traditional Offline RL methods, such as \textbf{CQL}~\cite{kumar2020conservative}, \textbf{IQL}~\cite{kostrikov2021iql}, \textbf{BCQ}~\cite{fujimoto2019off}, \textbf{TD3+BC}~\cite{fujimoto2021minimalist}, and \textbf{RvS}~\cite{emmons2021rvs}, mitigate distributional shift via conservatism constraints but are limited by unimodal assumptions.
To capture the full multimodal distribution of expert behavior~\cite{pearce2023imitating, pearce2022imitating}, \textbf{Diffusion Models (DMs)}~\cite{chi2025diffusion} have emerged as the state-of-the-art.
Variants cover diverse settings: \textbf{Diffusion-QL}~\cite{wang2023diffusion} and \textbf{DIPO}~\cite{yang2023policy} for offline RL, and \textbf{SfBC}~\cite{chen2023offline} for behavior cloning.
For hierarchical and constraint-aware planning, \textbf{Decision Diffuser}~\cite{ajay2023is} and \textbf{AdaptDiffuser}~\cite{liang2023adaptdiffuser} introduce inverse dynamics and goal conditioning, while \textbf{Decision Stacks}~\cite{zhao2024decision} decompose the policy into modular generative stacks.
\textbf{UniPi}~\cite{du2024learning} further extends this to video-based universal policies.
Similarly, \textbf{Normalizing Flows} offer bijective mapping for policy modeling, as seen in NF-Policy~\cite{ward2019improving} and Guided Flows~\cite{zheng2023guided}.

\vspace{6pt}
\noindent \textbf{Sequence Modeling: Autoregressive Transformers.}
Models like Decision Transformer (DT)~\cite{chen2021decision} reframe control as next-token prediction ($\tau = s_1, a_1, \dots$), leading to numerous extensions such as Trajectory Transformer~\cite{janner2021offline}, Online DT~\cite{zheng2022online}, and Bootstrapped DT~\cite{wang2022bootstrapped}. This paradigm scales exceptionally well with data size, serving as the backbone for Generalist Agents like Gato~\cite{reed2022generalist}, RT-1~\cite{brohan2022rt}, and multi-objective agents like PEDA~\cite{zhu2023scaling}.
However, standard autoregressive cloning is prone to compounding errors in open-loop generation and often struggles to extrapolate behavior beyond the training distribution.

To address these limitations and handle complex reasoning horizons, the field has expanded to leverage Large Language Models (LLMs) as High-Level Planners. Instead of outputting raw actions, frameworks like \textbf{SayCan}~\cite{ahn2022can} and \textbf{Code as Policies}~\cite{liang2023code} ground natural language instructions into executable primitives or Python code, enabling zero-shot robotic control. Prompting strategies such as \textbf{ReAct}~\cite{yao2022react} and \textbf{Inner Monologue}~\cite{huang2022inner} further introduce closed-loop feedback and self-correction mechanisms. In open-ended digital environments, autonomous agents like \textbf{Voyager}~\cite{wang2023voyager} and \textbf{GITM}~\cite{zhu2023ghost} demonstrate lifelong learning capabilities, while \textbf{Generative Agents}~\cite{park2023generative} utilize generative memory to simulate believable social behaviors.

\vspace{6pt}
\noindent \textbf{Synthesis and Model Selection.}
Selecting a generative controller requires trading off latency, mode coverage, and scalability. GANs and VAEs dominate latency-critical tasks (e.g., $>50$ Hz reactive control) but suffer from limited expressivity. Conversely, Diffusion Policies excel in high-fidelity offline imitation, prioritizing multimodal precision over inference speed. Finally, despite lacking the continuous precision of diffusion, Autoregressive Transformers leverage proven scaling laws to serve as the foundational architecture for large-scale generalist agents.

\subsection{Generative Models as Modeler (The Simulator)}
\label{subsec:modeler}

The Modeler approximates the transition dynamics $p(s_{t+1}|s_t, a_t)$ as defined in the factorization of Eq.~(\ref{eq:factorization}). By learning to synthesize environmental dynamics $P(s'|s,a)$ or counterfactual experiences, it serves as the physical prior that constrains trajectory inference. Unlike traditional augmentation methods like \textbf{S4RL}~\cite{sinha2022s4rl} or \textbf{RAD}~\cite{laskin2020reinforcement} which rely on heuristic perturbations, generative modelers capture the underlying data manifold~\cite{thomas2023integrating, ashuach2023multivi}.

\vspace{6pt}
\noindent \textbf{Latent Space Dynamics (RSSM \& VAEs).}
The concept of learning internal simulators traces back to the seminal \textbf{World Models}~\cite{ha2018world} and \textbf{PlaNet}~\cite{hafner2019learning}. While \textbf{MuZero}~\cite{schrittwieser2020mastering} pioneered value-equivalent planning without reconstructing observations, recent approaches like \textbf{VideoGPT}~\cite{yan2021videogpt} and \textbf{DayDreamer}~\cite{wu2022daydreamer} explicitly target high-fidelity visual synthesis. A cornerstone in this domain is the \textbf{Dreamer} family~\cite{hafner2019dream}, which utilizes Recurrent State Space Models (RSSM) to decouple deterministic history from stochastic transitions. By predicting future rewards and planning entirely within this compact latent space, these agents achieve unprecedented sample efficiency and fast inference for real-time Model Predictive Control (MPC). Extending this paradigm, VAE-based approaches~\cite{han2022selective, cho2022s2p} map high-dimensional states to structured spaces to bridge the pixel-to-control gap. However, a fundamental limitation of VAE-based dynamics is posterior collapse: the reconstruction objective often ignores complex, high-frequency visual details to focus on easily predictable features. Consequently, while latent modelers excel at core control, they may struggle in environments demanding precise spatial reasoning or fine-grained visual discrimination.

\vspace{6pt}
\noindent \textbf{Discrete Token Dynamics (Transformers).}
A rapidly emerging paradigm, exemplified by \textbf{IRIS}~\cite{micheli2022transformers} and \textbf{Genie}~\cite{bruce2024genie}, discretizes the world into visual tokens (via VQ-VAE) and models dynamics as autoregressive token prediction.
Recent works like ALPINE~\cite{wang2024alpine} and ARP~\cite{korenkevych2019autoregressive} extend this to action-conditioned prediction for long-term planning and exploration.
This approach leverages the scaling laws of Transformers, allowing world models to be trained on massive internet-scale video datasets. Unlike RSSMs which struggle with long-horizon consistency, Transformer-based modelers excel at maintaining coherence over long sequences, enabling Generative Interactive Environments where agents can train entirely inside a hallucinated world.

\vspace{6pt}
\noindent \textbf{High-Fidelity Observation Simulation (Diffusion, GANs \& Flows).}
To address the blurriness of VAEs, approaches leverage \textbf{Diffusion Models} (e.g., ROSIE~\cite{yu2023scaling}, GenAug~\cite{chen2023genaug}, MTDiff~\cite{he2024diffusion}) or \textbf{GANs} (e.g., EGAN~\cite{huang2017enhanced}) to synthesize photorealistic scenes and counterfactual scenarios.
Normalizing Flows (NFs) also contribute by modeling complex posterior distributions for efficient Bayesian inference~\cite{rezende2015variational, muller2019neural}.
These models excel at Sim-to-Real transfer and data augmentation by covering long-tail distributions. However, their high inference latency makes them largely unsuitable for real-time online planning loops, restricting their use primarily to offline data synthesis or low-frequency re-planning.


\vspace{6pt}
\noindent \textbf{Synthesis and Model Selection.}
The selection of a generative modeler depends on the role of simulation in the learning pipeline.
For online planning where thousands of imaginary trajectories must be evaluated per second, Latent Modelers (RSSM) remain the only viable option due to their compact state space.
For large-scale pre-training, Transformer-based Modelers offer a scalable path to learn general-purpose world models from diverse video data.
Finally, for offline robustness testing or generating synthetic training data for corner cases, Pixel-space Diffusion provides the visual fidelity to bridge the sim-to-real gap.

\subsection{Generative Models as Optimizer (The Planner)}
\label{subsec:optimizer}

The Optimizer serves as the inference engine responsible for solving the trajectory posterior $p(\tau | \mathcal{O}=1)$ derived in Eq.~(\ref{eq:factorization}). By effectively navigating high-dimensional density landscapes, these optimizers leverage generative priors to directly search for optimal trajectories $\tau^*$ or parameters $\theta^*$. Unlike feed-forward Controllers, Optimizers treat decision-making as an \textbf{iterative inference} or test-time sampling problem~\cite{qiao2019defending, lattimore2020learning}, strategically trading off computational budget for higher precision and long-horizon consistency.

\vspace{6pt}
\noindent \textbf{Trajectory In-painting (Diffuser).}
Approaches such as \textbf{Diffuser}~\cite{janner2022planning} reframe planning as a conditional in-painting task, where intermediate trajectories are generated via iterative denoising between a start state $s_0$ and a goal $s_g$. This framework has been extended by Decision Diffuser~\cite{ajay2023is} for constraint handling, AdaptDiffuser~\cite{liang2023adaptdiffuser} for goal-conditioned tasks, and Decision Stacks~\cite{zhao2024decision} for modular generation. \textbf{This formulation marks a fundamental departure} from traditional shooting methods, which rely on step-by-step dynamics rollouts and frequently suffer from compounding errors. By treating the entire trajectory $\tau$ as a single generative unit, these models ensure global temporal consistency and effectively mitigate the credit assignment challenges inherent in RL. \textbf{Nevertheless, the benefits of this global perspective are tempered by} a significant increase in inference latency, as the iterative denoising process ($O(K)$ steps) is computationally more demanding than traditional single-step reactive policies.

\vspace{6pt}
\noindent \textbf{Proportional Sampling (GFlowNets).}
\textbf{GFlowNets}~\cite{bengio2021gflownet, bengio2021flow} provide a rigorous framework for discrete and continuous optimization by learning to sample candidates proportional to the reward distribution: $\pi(x) \propto R(x)$. Recent advancements, including Stochastic GFN~\cite{pan2023stochastic}, GAFlowNets~\cite{pan2022generative}, and Continuous GFlowNets (CFlowNets)~\cite{li2023cflownets, lahlou2023theory}, have expanded its utility in complex control tasks. \textbf{The primary strength of this framework lies in} its inherent diversity-seeking mechanism, which renders GFlowNets superior to traditional MCMC or genetic algorithms in navigating multi-modal landscapes. While conventional optimizers often converge to a single local optimum (mode collapse), GFlowNets maintain coverage over multiple high-reward modes. \textbf{This property is particularly indispensable} for tasks such as scientific discovery or complex motion planning, where capturing a diverse set of high-quality solutions is as critical as finding the global optimum.

\vspace{6pt}
\noindent \textbf{Latent Space and Black-box Optimization.}
Generative models further facilitate efficient search by mapping rugged, high-dimensional optimization landscapes onto smooth latent manifolds. This strategy is exemplified by \textbf{VAEs} (e.g., CageBO~\cite{xing2023bayesian}, CVAE-Opt~\cite{hottung2021learning}) and \textbf{GANs} used in topology optimization~\cite{sim2021gans, he2020evolutionary}, as well as \textbf{Diffusion models} applied to black-box optimization like DDOM~\cite{krishnamoorthy2023diffusion} and DiffOPT~\cite{kong2024diffusion}. Autoregressive models, such as BONET~\cite{mashkaria2023generative} and TNPs~\cite{nguyen2022transformer}, similarly leverage pre-training to streamline the optimization process. \textbf{The core intuition behind this strategy is} the circumvention of local optima prevalent in raw parameter spaces. \textbf{Crucially, however, the efficacy of this approach remains strictly contingent upon} the fidelity and smoothness of the learned latent manifold; any discontinuities or poorly modeled regions in the latent space can inadvertently lead to sub-optimal or physically infeasible solutions.

\vspace{6pt}
\noindent \textbf{Synthesis and Model Selection.}
The optimal generative optimizer is determined by the topology of the solution space.
For continuous control tasks requiring long-horizon reasoning (e.g., maze navigation), Diffuser-style In-painting excels by treating time as a spatial dimension.
For discrete, combinatorial discovery tasks where the goal is to find a diverse set of high-performing candidates (e.g., drug discovery), GFlowNets are currently unmatched.
Finally, for black-box optimization problems where gradients are unavailable, Latent Space Optimization provides a differentiable surrogate landscape to accelerate convergence.

\subsection{Generative Models as Evaluator (The Critic)}
\label{subsec:evaluator}

The Evaluator approximates the optimality likelihood term $p(\mathcal{O}|\tau) \propto \exp(R(\tau))$ derived in Section~\ref{subsec:theory}, essentially serving as the grounding mechanism for the trajectory posterior. Generative models upgrade this role from simple scalar scoring to \textbf{distributional guidance} and \textbf{safety verification}.

\vspace{6pt}
\noindent \textbf{Energy-Based Guidance (EBMs).}
EBMs learn an unnormalized density function $E(s,a)$ which serves as a learnable cost function, often used in Inverse RL (e.g., EBIL~\cite{liu2021energy}, DEBMs~\cite{haarnoja2017reinforcement}) to capture expert reward structures. Importantly, unlike black-box reward models, EBMs are differentiable with respect to the input actions. This architectural trait allows planners to use the energy gradient $\nabla_a E(s,a)$ to optimize trajectories directly, proving particularly powerful for satisfying \textbf{soft constraints} during motion planning.

\vspace{6pt}
\noindent \textbf{Adversarial \& Preference Learning (Discriminators).}
In frameworks like GAIL~\cite{ho2016generative}, the Discriminator $D(s,a)$ acts as a surrogate reward signal. Modern extensions generalize this to learn from human preferences. Despite their effectiveness for Inverse RL, a major limitation is that these evaluators are inherently unstable due to the non-stationarity of the adversarial game. Specifically, if the generator outpaces the discriminator (reward hacking), the evaluation signal collapses.

\vspace{6pt}
\noindent \textbf{Density-Based Safety Monitors.}
Explicit density models (Normalizing Flows, VAEs) function as \textbf{OOD Detectors}. By computing the exact log-likelihood $\log p_\theta(s,a)$, they quantify the familiarity of a state. Consequently, this density estimation serves as a fundamental component for \textbf{Safe Deployment}. If a proposed action falls into a low-density region (Out-of-Distribution), the Evaluator can veto the action or penalize the reward. This epistemic uncertainty estimation is therefore the primary defense against hallucinated or dangerous behaviors in open-world environments.

\vspace{6pt}
\noindent \textbf{Synthesis and Model Selection.}
The choice of Evaluator is dictated by the availability of supervision signals.
When expert demonstrations are available but no reward function exists, Discriminators (Adversarial Learning) are essential for extracting surrogate rewards.
When explicit constraints must be satisfied, EBMs offer the geometric gradients needed for optimization.
Crucially, for safety-critical deployment, incorporating a Density-Based Monitor is not optional but necessary to filter unreliable model outputs.

\section{Applications and Safety Analysis}
\label{sec:applications}

While generative models have demonstrated transformative capabilities across diverse domains, their deployment in high-consequence decision-making introduces non-trivial safety risks. In this section, we move beyond algorithmic details to critically analyze the \textbf{robustness, safety, and misuse challenges} inherent to each domain. We focus on three high-impact areas where the tension between generative expressivity and reliability is most acute. A summary of these risks and mitigation strategies is provided in Table~\ref{tab:app_risks}.

\begin{table*}[t]
\centering
\caption{\textbf{Summary of Generative Applications, Systemic Risks, and Mitigation Strategies.} We map each domain's primary utility to our proposed functional taxonomy (in parentheses), highlighting the specific vulnerabilities induced by generative mechanisms and current mitigation frontiers.}
\label{tab:app_risks}
\renewcommand{\arraystretch}{1.3}
\resizebox{\textwidth}{!}{%
\begin{tabular}{l|l|l|l|l}
\toprule
\textbf{Domain} & \textbf{Core Generative Utility (Role)} & \textbf{Key Benefit} & \textbf{Systemic Risk} & \textbf{Mitigation Strategy} \\
\midrule
\multirow{2}{*}{\textbf{Embodied AI \& Robotics}} 
& World Simulation (\textbf{Modeler}) & Infinite Synthetic Data & \textbf{Physics Hallucination} (Unreal dynamics) & Neuro-symbolic Physics; Sim-to-Real \\
& Generalist Policies (\textbf{Controller}) & Multimodal Behavior & \textbf{High-Confidence Errors} (OOD shift) & Conformal Prediction~\cite{ren2023knowno}; Uncertainty Bounds \\
\midrule
\multirow{2}{*}{\textbf{Autonomous Driving}} 
& Corner Case Synthesis (\textbf{Modeler}) & Long-tail Coverage & \textbf{Sensor/Domain Shift} & Adversarial Training; Generative Augmentation \\
& End-to-End Planning (\textbf{Optimizer}) & Joint Occupancy & \textbf{Semantic Adversarial Attacks} & Hierarchical Safeguards (e.g., RSS) \\
\midrule
\multirow{2}{*}{\textbf{Scientific Discovery}} 
& Structure Optimization (\textbf{Optimizer}) & Diverse Candidate Search & \textbf{Proxy Exploitation} (Invalid structures) & Latent Sanitization; In-loop Verification \\
& Biochemical Design (\textbf{Evaluator}) & Differentiable Guidance & \textbf{Dual-Use / Biosecurity} (Toxin generation) & Machine Unlearning; Alignment Guardrails \\
\bottomrule
\end{tabular}%
}
\end{table*}

\subsection{Embodied AI and Robotics}
\label{subsec:robotics}

In the domain of Embodied AI, generative models are fundamentally reshaping the learning paradigm, advancing beyond classical control methods~\cite{duan2016benchmarking, levine2016end, bailo2018optimal}. They act both as \textbf{infinite data engines} that mitigate data scarcity and \textbf{generalist policy priors} that enable robust generalization, as seen in foundation agents like Gato~\cite{reed2022generalist} and Multi-Game Transformers~\cite{lee2022multigame, li2024unbounded}.

\vspace{6pt}
\noindent \textbf{Generative Simulation and The Reality Gap.}
The scarcity of diverse real-world interaction data remains the primary bottleneck for robot learning. Generative modelers address this by synthesizing large-scale synthetic environments. 
To bridge the visual sim-to-real gap, early works relied on \textbf{Domain Randomization}~\cite{tobin2017domain, peng2018sim} and feature-level adaptation~\cite{hoffman2018cycada}. 
Generative approaches extend this by semantically modifying simulation assets (e.g., \textbf{GenAug}~\cite{chen2023genaug}) or synthesizing realistic textures via GANs (e.g., \textbf{RetinaGAN}~\cite{ho2021retinagan})~\cite{hofer2021sim2real, chenunderstanding, wang2018deep}.
More recently, foundation models have enabled \textbf{automated curriculum generation}; systems like \textbf{RoboGen}~\cite{wang2023robogen} leverage LLMs and generative models to autonomously propose tasks and synthesize demonstration trajectories.

\vspace{6pt}
\noindent \textbf{Multimodal Policy Learning and Safety.}
On the control side, the field has witnessed a paradigm shift from unimodal Gaussian policies to \textbf{generative controllers}, predominantly driven by Diffusion Models~\cite{zhang2023diffusion, malkin2022trajectory}. Approaches like \textbf{Diffusion Policy}~\cite{chi2025diffusion} model the policy as a conditional denoising process. This formulation captures the highly multi-modal action distributions inherent in human demonstrations (e.g., bypassing an obstacle from either the left or right), which standard MSE-based cloning fails to represent. This expressivity has culminated in generalist Vision-Language-Action (VLA) policies, such as \textbf{Octo}~\cite{team2024octo} and \textbf{OpenVLA}~\cite{kim2024openvla}, extending even to complex locomotion~\cite{margolis2024rapid}.

Despite their success, the stochastic nature of generative policies complicates \textbf{safety verification}~\cite{hu2022provable}. A primary vulnerability is \textbf{high-confidence hallucination} during distributional shifts, where diffusion policies may generate visually coherent yet hazardous trajectories without intrinsic uncertainty signaling. Mitigating this requires integrating rigorous uncertainty quantification, such as \textbf{conformal prediction}~\cite{ren2023knowno} for statistical safety bounds, to detect and abort unsafe executions.

\subsection{Autonomous Driving (AD)}
\label{subsec:driving}

In autonomous driving, the long-tail distribution of safety-critical scenarios necessitates a transition from traditional log-replay to \textbf{generative simulation}, and from modular pipelines to \textbf{end-to-end generative planning}~\cite{ghosh2016sad, chen2018parallel, wang2021decision}.

\vspace{6pt}
\noindent \textbf{Corner Case Synthesis and Domain Fidelity.}
Naturalistic driving logs are dominated by mundane cruising, leaving critical corner cases underrepresented. Generative modelers address this through \textbf{controllable scene synthesis}. Foundation models like \textbf{MagicDrive}~\cite{gao2024magicdrive} leverage layout-conditioned diffusion to synthesize photorealistic, multi-view sensor data, enhancing perception tasks~\cite{marathe2023wedge, arnelid2019recurrent}. Furthermore, world models like \textbf{Drive-WM}~\cite{wang2024drivewm} and TrafficGen~\cite{feng2023trafficgen} facilitate counterfactual safety testing by simulating potential future trajectories and edge cases.

However, a persistent gap remains in \textbf{sensor-realistic consistency}. Generative simulators often struggle to preserve high-frequency sensor characteristics, creating a domain shift that degrades planner performance. Moreover, generative scene reconstruction is vulnerable to \textbf{semantic adversarial attacks}, where imperceptible perturbations in latent spaces can mislead planners into predicting hazardous maneuvers.

\begin{figure}[b]
\centering
\includegraphics[width=\columnwidth]{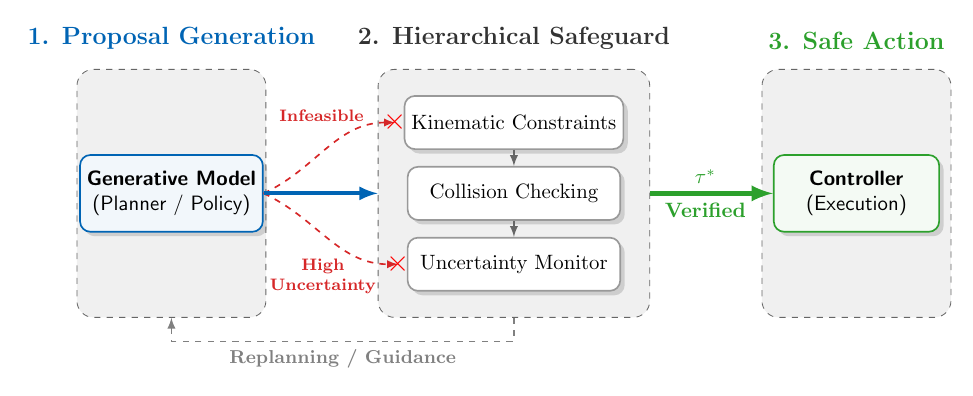}
\caption{\textbf{Hierarchical Safeguard System for Generative Decision-Making.} 
To mitigate the stochastic risks of generative models (e.g., hallucinations or constraint violations), the system employs a \textit{Generate-then-Filter} paradigm. 
(1) The \textbf{Generative Model} acts as a creative proposal distribution. 
(2) A deterministic \textbf{Safeguard Module} (e.g., RSS for driving or Collision Checkers for robotics) filters out infeasible actions. 
(3) Only verified trajectories are executed. This architecture decouples expressivity from safety assurance.}
\label{fig:safety_shield}
\end{figure}

\vspace{6pt}
\noindent \textbf{Generative Planning and Hierarchical Safeguards.}
The field is transitioning towards \textbf{end-to-end generative planning}, with systems like \textbf{UniAD}~\cite{hu2023planning} casting decision-making as a joint occupancy and ego-motion prediction task. Nevertheless, the black-box nature of generative models hinders \textbf{certifiable safety}. A model may output a high-likelihood trajectory that violates hard constraints (e.g., crossing lane boundaries). To align with safety standards like ISO 26262, deployment requires a \textbf{hierarchical safeguard system} (Figure~\ref{fig:safety_shield}). In this hybrid architecture~\cite{huang2021learning}, the generative model acts as a creative proposal distribution, while a safety filter grounded in formal logic (e.g., RSS~\cite{shalev2017formal}) or conformal prediction (e.g., PlanCP~\cite{sun2024conformal}) actively rejects unsafe actions.

\subsection{Scientific Discovery \& Material Design}
\label{subsec:design}

In biochemical discovery and combinatorial optimization, generative models shift the focus from behavior imitation to efficient combinatorial search~\cite{bengio2021machine, zhang2023survey}. By learning the manifold of valid structures, these models act as high-dimensional optimizers that balance sample fidelity with structural diversity for tasks ranging from graph generation~\cite{li2018multi, grover2019graphite} to neural architecture search~\cite{liu2023gflowout}.

\vspace{6pt}
\noindent \textbf{From Engineering Design to Scientific Discovery.}
Generative optimization reformulates decision-making as probabilistic sampling. Optimizers like \textbf{RFdiffusion}~\cite{watson2023novo} and \textbf{DiffDock}~\cite{corso2023diffdock} navigate complex chemical manifolds by sampling candidates proportional to their biological utility. 
This paradigm has been notably advanced by \textbf{AlphaFold 3}~\cite{abramson2024accurate} for biomolecular interaction prediction, and by geometric generators like \textbf{GeoDiff}~\cite{xu2022geodiff} and \textbf{Equivariant Diffusion}~\cite{hoogeboom2022equivariant}.
Similar generative principles apply to human kinematic synthesis (\textbf{MotionDiffuse}~\cite{zhang2024motiondiffuse}).
In the discrete domain, employing \textbf{GFlowNets}~\cite{bengio2021gflownet} to explore the posterior of optimal structures facilitates the discovery of novel protein backbones distinct from existing datasets~\cite{jain2022biological, kim2023local}. Research in prompt-based optimization~\cite{xin2022rethinking, deng2022rlprompt, zhang2023tempera} further highlights their versatility in handling diverse conditioning constraints.

\vspace{6pt}
\noindent \textbf{Reliability Gaps and Biosecurity Risks.}
Despite their expressive power, generative optimizers are susceptible to \textbf{proxy exploitation}, a manifestation of Goodhart’s Law where models hack the inaccuracies of learned surrogate reward functions. By optimizing against imperfect proxies, generators often produce chemically invalid or structurally unstable candidates. More critically, their dual-use potential poses significant \textbf{biosecurity risks}, as mechanisms designed for therapeutics can be repurposed to generate toxic pathogens. To mitigate these threats, the field is actively developing \textbf{latent space sanitization}~\cite{schwalbe2024predictable} and human-in-the-loop verification to constrain exploration to ethical and physically plausible regions.

\section{Open Challenges and Future Directions}
\label{sec:challenges}

While generative models have initiated a paradigm shift in decision-making, bridging the gap between current capabilities and deployment-ready agents requires addressing several challenges. We highlight four pivotal directions for the next generation of \textbf{generalist physical intelligence}.

\subsection{Foundation Models for Physical Intelligence}
\label{subsec:foundation}

The field is witnessing a paradigm shift from domain-specific controllers to \textbf{Physical Foundation Models (PFMs)} that natively model the continuous dynamics of the physical world~\cite{intelligence2025pi0}. 
Unlike early Vision-Language-Action models (e.g., RT-2~\cite{zitkovich2023rt, brohan2022rt}, OpenVLA~\cite{kim2024openvla}, and CogACT~\cite{li2024cogact}) that treated physical interaction as a discrete text completion problem, next-generation PFMs aim to bridge high-level semantic reasoning with low-level physical execution across diverse platforms.
This evolution centers on three critical architectural dimensions: efficient sequence processing, predictive world modeling, and continuous generative action.

\vspace{6pt}
\noindent \textbf{Efficient Sequence Backbones.}
Standard Transformers suffer from quadratic computational complexity $O(N^2)$, creating a bottleneck for processing dense, high-frequency sensorimotor streams.
To mitigate this, emerging research investigates State-Space Models (SSMs), such as \textbf{Mamba}~\cite{gu2024mamba}, as linear-complexity alternatives.
In perception, architectures like \textbf{Vision Mamba}~\cite{zhu2024vision} demonstrate that visual SSMs can process high-resolution temporal inputs with significantly lower overhead.
Extending this to decision-making, frameworks like \textbf{Cobra}~\cite{zhao2025cobra} adapt SSMs to multi-modal reasoning. By compressing history into fixed-size recurrent states, these backbones efficiently capture the long-horizon dependencies crucial for complex manipulation tasks.

\vspace{6pt}
\noindent \textbf{General-Purpose World Models.}
Learning robust internal models of physical dynamics is fundamental to embodied intelligence.
One prominent paradigm focuses on Latent Prediction (e.g., \textbf{V-JEPA}~\cite{bardes2024revisiting}), while generative video-based world models like \textbf{Genie}~\cite{bruce2024genie} treat physical interaction as a controllable generative process.
The recent \textbf{NVIDIA Cosmos}~\cite{agarwal2025cosmos} scales this approach, providing a world foundation model that adheres to conservation laws.
By serving as scalable engines for both data synthesis and counterfactual planning, these models provide a unified substrate for agents to evaluate potential futures across diverse physical domains.

\vspace{6pt}
\noindent \textbf{Generative Action Modeling.}
A critical limitation of early foundation models lies in quantization errors. To address this, the field is shifting towards continuous generative modeling.
Diffusion-based approaches, such as \textbf{Diffusion Policy}~\cite{chi2025diffusion} and web-scale robotic diffusion models~\cite{kapelyukh2023dall, yu2023scaling, gupta2025pre}, treat trajectory planning as a probabilistic denoising process.
Recent advancements, including \textbf{DP3}~\cite{ze20243d} for spatial generalization and \textbf{Consistency Policy}~\cite{prasad2024consistency} for high-frequency control, have specialized these models for physical agency.
Integrating these into unified backbones like \textbf{Pi0}~\cite{intelligence2025pi0} enables agents to bridge abstract semantic goals with high-fidelity, continuous motor execution.

\subsection{Real-Time Inference and Sampling Efficiency}
\label{subsec:inference}

A critical bottleneck preventing deployment is \textit{inference latency}. Bridging the frequency gap between Hertz-level generation and Kilohertz-level control necessitates a transition to accelerated generation architectures.

\vspace{6pt}
\noindent \textbf{Distillation and Speculative Execution.}
Techniques like \textbf{Consistency Models}~\cite{song2023consistency} learn to map arbitrary noise states directly to the solution manifold in a single step, distilling the knowledge of slow diffusion teachers. In parallel, \textbf{Speculative Decoding}~\cite{leviathan2023fast} offers a runtime acceleration strategy for autoregressive planners, trading parallel computation for reduced latency.

\vspace{6pt}
\noindent \textbf{Flow Matching and Straight Paths.}
Beyond distillation, a more fundamental approach lies in reformulating the generative ODE. \textbf{Rectified Flow}~\cite{liu2023flow} and Conditional Flow Matching~\cite{lipman2022flow} enforce straight generation trajectories in the probability flow, minimizing transport cost. This enables high-quality sample generation with as few as 1--2 Euler steps, promising SOTA efficiency for real-time planning.

\vspace{6pt}
\noindent \textbf{Action Chunking and Cognitive Decoupling.}
At the deployment level, engineering paradigms like \textbf{Action Chunking}~\cite{zhao2023learning} effectively mask inference latency by generating and executing multi-step open-loop macro-actions, often combined with temporal ensembling for smoothness. 
Concurrently, inspired by biological cognition, hierarchical frameworks like \textbf{SwiftSage}~\cite{lin2023swiftsage} decouple decision-making into deliberative planning (System 2) and reflexive execution (System 1), reconciling the latency of foundation models with real-time sensorimotor responsiveness.

\subsection{Trustworthy AI: Safety, Alignment, and Unlearning}
\label{subsec:trustworthy}

The transition from disembodied agents to embodied actors mandates that next-generation foundation models be verifiable, behaviorally aligned, and corrigible.

\vspace{6pt}
\noindent \textbf{Certifiable Constrained Generation.}
To address the lack of safety guarantees, research is embedding rigorous constraints into the sampling process.
Approaches like \textbf{SafeDiffuser}~\cite{he2025safediffuser} integrate Control Barrier Functions (CBFs) into the reverse diffusion SDE to enforce invariant sets.
Statistically, applying conformal prediction to physical foundation models (e.g., \textbf{KnowNo}~\cite{ren2023knowno}) allows agents to quantify uncertainty with finite-sample guarantees, deferring actions when confidence is low.

\vspace{6pt}
\noindent \textbf{Safety-Aware Preference Alignment.}
Embodied control requires safety-aware preference optimization beyond standard RLHF~\cite{ouyang2022training}.
Emerging paradigms adapt constitutional AI principles to physical agents (e.g., \textbf{AutoRT}~\cite{ahn2024autort}), where models are fine-tuned on compliance with operational constitutions.
Furthermore, optimization techniques like \textbf{Direct Preference Optimization (DPO)}~\cite{rafailov2023direct} and \textbf{LIMA}~\cite{zhou2023lima} offer stable alignment objectives without separate reward models.
Recent efforts also explore using LVLMs as independent safety verifiers~\cite{huang2023voxposer} to prune unsafe actions.

\vspace{6pt}
\noindent \textbf{Machine Unlearning for Anti-Misuse.}
The risk of dual-use creates a need for post-training safety measures.
\textbf{Machine unlearning}~\cite{cao2015towards} aims to erase dangerous knowledge without damaging overall performance.
Unlike unlearning in LLMs, unlearning in physical policies requires erasing specific behavioral primitives (e.g., synthesizing toxins) while preserving basic motor skills, which remains an open challenge for future research~\cite{chundawat2023can}.

\subsection{Theoretical Foundations}
\label{subsec:theory}

Bridging experimental success with rigorous guarantees is crucial for safety-critical deployment.

\vspace{6pt}
\noindent \textbf{Causal Identifiability.}
Current world models relying on observational data often capture spurious correlations (causal confusion) rather than true physical mechanisms~\cite{de2019causal}.
Future theory must integrate causal discovery into generative modeling~\cite{scholkopf2021toward}, transitioning from correlational simulators to \textbf{Causal World Models} that support valid counterfactual reasoning and intervention.

\vspace{6pt}
\noindent \textbf{Generalization and Mode Coverage.}
We hypothesize that the superior performance of generative policies stems from their ability to cover the full distribution of optimal behaviors (Mode Coverage).
Rigorous work is needed to derive sample complexity bounds~\cite{chen2023sampling} that explicitly link the diversity of the generative posterior to the policy's success rate in OOD scenarios, particularly in high-dimensional continuous spaces.

\section{Conclusion}
\label{sec:conclusion}

In this survey, we have systematized the rapidly evolving landscape of generative models in decision-making. 
Departing from conventional architecture-centric taxonomies, we established a unified framework grounded in the perspective of Control as Inference. 
By mathematically factorizing the probabilistic control loop, we delineated four distinct functional roles for generative mechanisms: acting as Controllers that amortize policy inference, Modelers that enforce dynamics priors, Optimizers that refine trajectories via iterative sampling, and Evaluators that provide dense likelihood guidance.

Our analysis reveals a fundamental transition in the field: from the scalar reward maximization of classical Reinforcement Learning to high-fidelity distribution matching. 
Unlike standard policies that often collapse into brittle point estimates, generative models approximate the full trajectory posterior. 
This expressivity is the cornerstone for addressing modern challenges, including multi-modal imitation, robust offline learning from suboptimal data, and open-ended exploration in high-dimensional spaces.

Looking ahead, the convergence of generative AI and physical systems signals the dawn of Generalist Physical Intelligence. 
However, bridging the gap between generative simulation and physical execution requires addressing critical challenges, including inference efficiency, safety verification, and causal reasoning. 
We envision next-generation agents that go beyond merely hallucinating plausible futures to effectively realizing them in the physical world.
    
\ifCLASSOPTIONcaptionsoff
  \newpage
\fi

\bibliographystyle{IEEEtran}
\bibliography{IEEEabrv,refs}

\end{document}